\documentclass[10pt,twocolumn,letterpaper]{article}

\usepackage{iccv}
\usepackage{times}
\usepackage{epsfig}
\usepackage{graphicx}
\usepackage{amsmath}
\usepackage{amssymb}
\usepackage{booktabs}
\usepackage{array}
\usepackage{bm}
\usepackage{color, colortbl}
\usepackage{abdalmageed_style}

\usepackage[pagebackref=true,breaklinks=true,letterpaper=true,colorlinks,bookmarks=false]{hyperref}
\usepackage[capitalise]{cleveref}
\iccvfinalcopy % *** Uncomment this line for the final submission

\def\iccvPaperID{9899} % *** Enter the ICCV Paper ID here

\ificcvfinal\fi

\newcommand{\PreserveBackslash}[1]{\let\temp=\\#1\let\\=\temp}
\newcolumntype{C}[1]{>{\PreserveBackslash\centering}m{#1}}
\newcolumntype{R}[1]{>{\PreserveBackslash\raggedleft}m{#1}}
\newcolumntype{L}[1]{>{\PreserveBackslash\raggedright}m{#1}}
\newcommand{\netname}[1]{\textit{\textbf{#1}}}
\newcolumntype{g}{>{\columncolor{Gray}}c}
\def\simfullname{Spatial Information Module\,}
\def\sim{SIM\,} %spatial module name, spatial information module?
\def\modelfullname{Spatial-information Incorporated Generative Network\,}
\def\modelname{SIGN} % Spatial-information incorporated generative network?

\begin{document}
\setlength{\belowdisplayskip}{0pt} \setlength{\belowdisplayshortskip}{0pt}
\setlength{\abovedisplayskip}{5pt} \setlength{\abovedisplayshortskip}{2pt}
%%%%%%%%% TITLE
\title{SIGN: Spatial-information Incorporated Generative Network \\ for Generalized Zero-shot Semantic Segmentation}

\author{Jiaxin Cheng, Soumyaroop Nandi, Prem Natarajan, Wael Abd-Almageed\\
USC Information Sciences Institute, Marina del Rey, CA, USA\\
{\tt\small \{chengjia,soumyarn,pnataraj,wamageed\}@isi.edu}}

\maketitle
\ificcvfinal\thispagestyle{empty}\fi

%%%%%%%%% ABSTRACT
\begin{abstract}
    Unlike conventional zero-shot classification, zero-shot semantic segmentation predicts a class label at the pixel level instead of the image level. When solving zero-shot semantic segmentation problems, the need for pixel-level prediction with surrounding context motivates us to incorporate spatial information using positional encoding. We improve standard positional encoding by introducing the concept of Relative Positional Encoding, which integrates spatial information at the feature level and can handle arbitrary image sizes. Furthermore, while self-training is widely used in zero-shot semantic segmentation to generate pseudo-labels, we propose a new knowledge-distillation-inspired self-training strategy, namely Annealed Self-Training, which can automatically assign different importance to pseudo-labels to improve performance. We systematically study the proposed Relative Positional Encoding and Annealed Self-Training in a comprehensive experimental evaluation, and our empirical results confirm the effectiveness of our method on three benchmark datasets. 
\end{abstract}
% . The need for pixel-level labeling

%%%%%%%%% BODY TEXT
% \vspace{-10px}
\section{Introduction}
\label{sec:intro}
Zero-shot learning (ZSL) solves the task of learning in the absence of training data of that task (\eg, recognizing unseen classes). It has been widely adopted in classic computer vision problems, such as classification  \cite{akata2015label,frome2013devise,akata2015evaluation,romera2015embarrassingly,socher2013zero,xian2016latent,zhang2015zero,norouzi2013zero,xian2018feature,chen2018zero,mishra2018generative,felix2018multi,long2017zero,morgado2017semantically,guo2017zero,kampffmeyer2019rethinking,wang2018zero} and object detection  \cite{rahman2018polarity,al2015transfer,fu2015zero,demirel2018zero,rahman2018zero,bansal2018zero}. The key challenge in ZSL based tasks is to make the underlying model capable of recognizing classes that had not been seen during training. Earlier work focused on learning a joint embedding between seen and unseen classes \cite{akata2015label,frome2013devise,akata2015evaluation,romera2015embarrassingly,socher2013zero,xian2016latent,zhang2015zero,norouzi2013zero}. In recent works, knowledge and generative-based  methods have gained more prominence. Knowledge-based methods \cite{morgado2017semantically,guo2017zero,kampffmeyer2019rethinking,wang2018zero} use structured knowledge learned in another domain (\eg, natural language \cite{miller1995wordnet}, knowledge graph\cite{liu2004conceptnet}, \etc) as constraints \cite{guo2017zero,kampffmeyer2019rethinking} to transfer the learned information to unseen categories. Generative methods use attributes \cite{felix2018multi,chen2018zero}, natural language \cite{ramesh2021zero} or word embeddings \cite{felix2018multi,chen2017rethinking} as priors to generate synthetic features for unseen categories. 

\begin{figure}
    \centering
    \includegraphics[width=\linewidth]{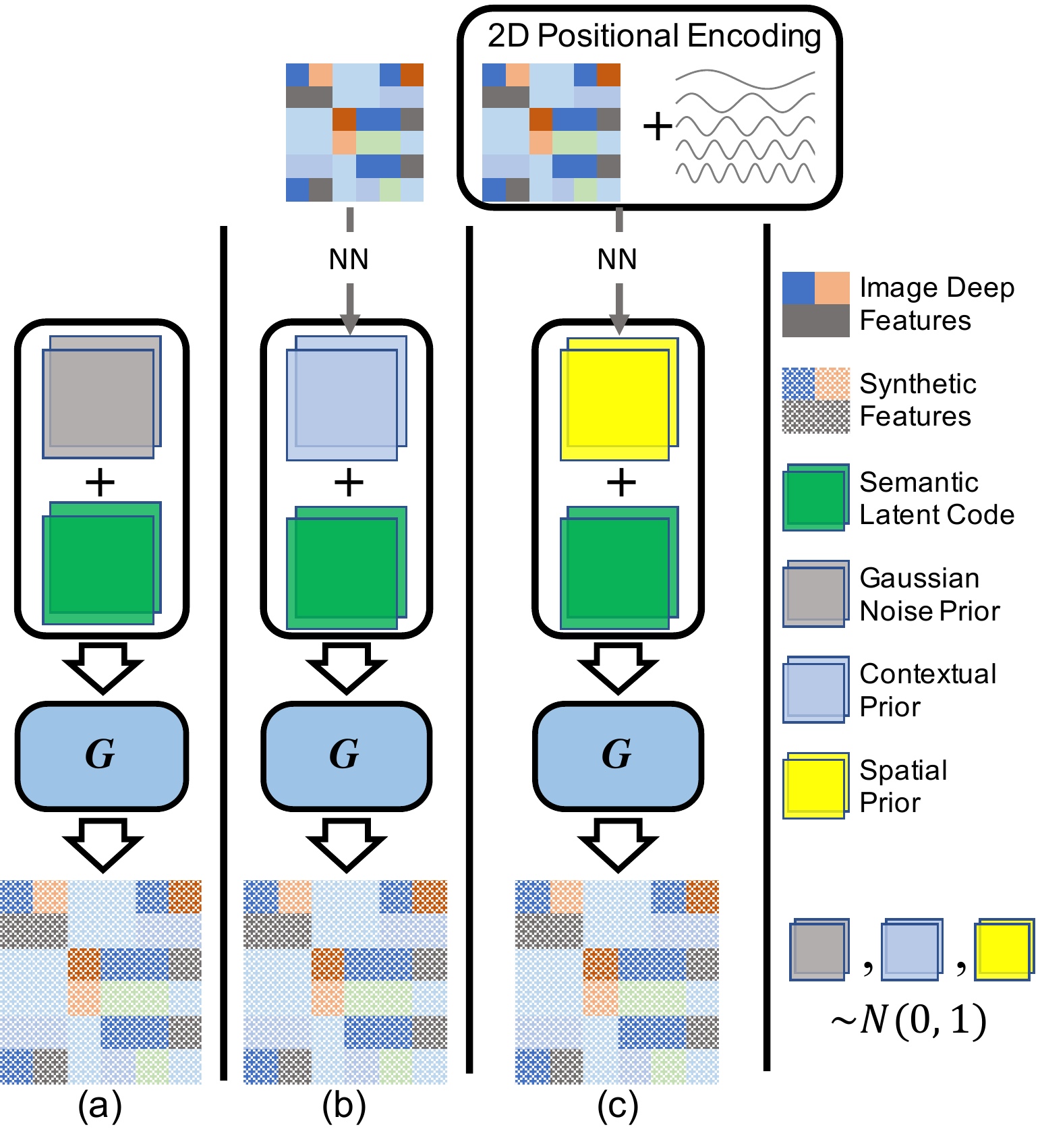}
    \caption{Latent code used in generating synthetic features consists of two parts: i) semantic word embeddings and ii) a normal distributed prior. The normal distributed prior can be: (a) Gaussian noise \cite{zs3}; (b) Context-aware prior \cite{cagnet}; (c) Our context-aware and space-aware priors.}
    \label{fig.latent_code}
    % \vspace{-10px}
\end{figure}

In this work, we investigate the zero-shot semantic segmentation problem, which is less studied than other zero-shot computer vision problems \cite{spnet,zs3,kato2019zero,cagnet}.
Generative methods have been widely adopted in zero-shot semantic segmentation problem. Bucher \etal \cite{zs3} proposed ZS3Net, in which they leverage Generative Moment Matching Networks \cite{li2015generative} to generate synthetic features for unseen categories by using word2vec \cite{word2vec} embeddings and random noise as prior (\cref{fig.latent_code}(a)). The use of random noise prevents the model from collapsing, which occurs due to lack of feature variety \cite{zhu2017unpaired,goodfellow2014generative}. Li \etal \cite{csrl} extended ZS3Net \cite{zs3} by adding a structural relation loss which constrains the generated synthetic features to have a similar structure as the semantic word embedding space. Gu \etal \cite{cagnet} suggested using a context-aware normal distributed prior when generating synthetic features instead of random noise (\cref{fig.latent_code}(b)). 

We propose to incorporate spatial information to improve the performance of the zero-shot semantic segmentation problem. Our motivation arises from the assumption that knowing a pixel's location may help semantic segmentation because it is a 2D prediction task. Incorporating spatial information in computer vision problems has recently attracted the attention of the community.  In image classification, \cite{dosovitskiy2020image} slices the input image into nine patches and adds a positional vector for each patch to indicate the patches' location. Other methods leverage \cite{xu2019spatial}  the relative position of objects through a space-aware knowledge graph for object detection. Zhang \etal \cite{zhang2019co} counted the co-occurrence of features to learn spatial invariant representation for semantic segmentation. However,  to the best of our knowledge, spatial information has not been widely studied in previous zero-shot learning research. In this work, we propose to exploit spatial information by using Positional Encoding~\cite{vaswani2017attention}, as shown in \cref{fig.latent_code}(c). Positional Encoding generates a positional vector that indicates the position of a pixel in the image. Previous work \cite{dosovitskiy2020image} divided the input images into fixed number of patches and appended positional embeddings on the images. However, in our case, dividing the input image into small patches is incompatible because the semantic segmentation problem requires the entire image as input. We mitigate this shortcoming by incorporating spatial information into the image features and propose Relative Positional Encoding to handle varying input image sizes. 

In the zero-shot learning problem, unlabeled samples, including unseen classes, are sometimes available. Trained models can annotate unlabeled samples to obtain additional training data \cite{zhu2005semi} and fine-tune the models with pseudo-annotated data. Such a training strategy is called self-training. In zero-shot semantic segmentation, self-training annotates pixel-level pseudo labels \cite{zs3,cagnet}. To reduce the number of unreliable pseudo labels, \cite{zs3} ranks the confidence score (\ie the probability of classes after \textit{softmax} function) of pseudo labels and uses only the most confident 75\% pseudo labels. \cite{cagnet} eliminates pseudo labels if the confidence score is below a certain threshold. 

This work proposes a knowledge distillation-inspired \cite{hinton2015distilling} self-training strategy, namely \emph{Annealed Self-Training} (AST), to generate better pseudo-annotations for self-training. Knowledge distillation is widely used in the form of teacher-student learning \cite{hinton2015distilling,mirzadeh2020improved}, where the student network is trained on the soft labels from the teacher network in addition to (one-hot) hard labels since soft labels have more information due to high entropy \cite{hinton2015distilling}. In the zero-shot semantic segmentation problem, previous methods \cite{zs3,cagnet} set a threshold to eliminate pseudo labels from the low confidence scores, while assigning the same loss weights to the remaining pseudo labels. This is similar to using hard labels in teacher-student learning. However, it is hard to ensure that the threshold can generalize for each sample, and again low confidence pseudo labels may also contain some useful information. To avoid the shortcomings of setting a threshold and assigning the same loss weights to pseudo labels, in AST, we use pseudo-annotations of all unlabeled pixels while re-weighing their importance according to their confidence score. We leverage the annealed \textit{softmax} \cite{hinton2015distilling} function to normalize the pseudo labels' weights and control their relative importance by adjusting the \emph{annealed temperature} in the \textit{softmax} function. 

We make the following contributions in this paper. Firstly, we introduce the Spatial Information Module to incorporate spatial information in semantic segmentation using a novel Relative Positional Encoding  (RPE) scheme. Compared to previous work \cite{dosovitskiy2020image}, RPE does not need patch-sliced input and can handle varying image sizes. Secondly, we propose a knowledge distillation-inspired self-training strategy, namely Annealed Self-Training (AST). AST generates pseudo-annotations for unlabeled samples and adjusts their importance during self-training with a tunable \emph{annealed temperature}. Finally, we evaluate the performance on three benchmark datasets and conduct extensive ablation experiments to demonstrate the effectiveness of our method.

% \vspace{-1px}
\section{Related Work}
% \vspace{-5px}

\begin{figure*}[t]
    \centering
    \includegraphics[width=\linewidth]{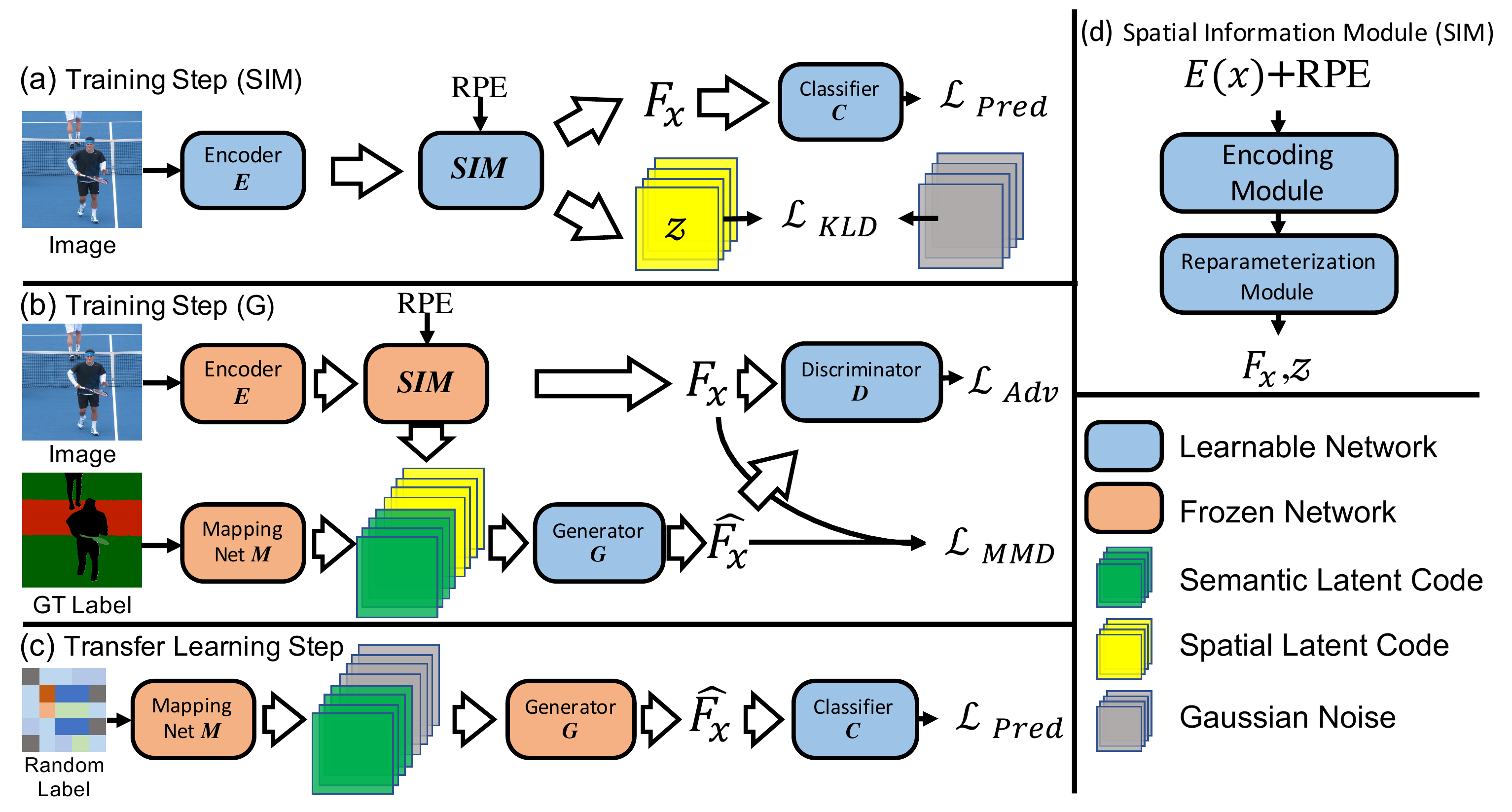}
    \caption{Our model consists of five learnable networks (\netname{E}, \netname{G}, \netname{C}, \netname{D}, \netname{SIM}) and one unlearnable network (\netname{M}). \netname{E} is the feature encoder CNN using Deeplab-v2 architecture. Generator \netname{G} generate synthetic features for unseen classes which can deceive discriminator \netname{D}. Classifier \netname{C} is trained on seen categories during Training Step and synthetic unseen categories during Transfer Learning Step. \netname{\sim} encodes spatial information into deep features through Relative Positional Encoding. Mapping network \netname{M} converts ground truth into semantic latent code.}
    \label{fig.network_architecture}
    \vspace{-15pt}
\end{figure*}

\noindent\textbf{Zero-shot Learning}
Without loss of generality, approaches to zero-shot classification can be categorized into three families --- joint embedding, generative and knowledge-based methods. Earlier works focused on linear embedding  \cite{akata2015label,frome2013devise,akata2015evaluation,romera2015embarrassingly}, non-linear embedding \cite{socher2013zero,xian2016latent}, and hybrid embedding  \cite{zhang2015zero,norouzi2013zero} methods. In the embedding-based methods, the basic idea is to learn encodings for images and attributes (\eg, description) and maximize a linear/non-linear score between matched pairs. Generative  \cite{xian2018feature,chen2018zero,mishra2018generative,felix2018multi,long2017zero} and knowledge-based \cite{morgado2017semantically,guo2017zero,kampffmeyer2019rethinking,wang2018zero} approaches have recently become more popular. Generative method use attributes to create synthetic images with a generative model (\eg, generative adversarial network \cite{goodfellow2014generative} or a conditional variational autoencoder \cite{sohn2015learning}) and then train a classifier based on seen and synthetic unseen categories. Knowledge-based method often use structured knowledge as constraints  \cite{guo2017zero,kampffmeyer2019rethinking,wang2018zero} of relationships between classes and employ graph networks, \eg, Graph Convolutional Network  \cite{kipf2016semi}, to generalize learned information from seen categories to unseen ones. 

\noindent\textbf{Zero-shot Semantic Segmentation}
Bucher \etal \cite{zs3} use Generative Moment Matching Networks \cite{li2015generative} to create synthetic features for unseen categories and train a classifier based on the union of the features of seen categories and synthetic features. Xian \etal \cite{spnet} replace the last layer of a classifier with embeddings from word2vec \cite{word2vec} and introduce a calibration mechanism to adjust the class probability on seen categories due to imbalanced confidence on seen and unseen classes. Kato \etal \cite{kato2019zero} used a variational mapping for binary zero-shot semantic segmentation. They create a two-branched -- conditioning and segmentation -- network. The conditioning branch takes embeddings of unseen classes and maps them to classification layers in the segmentation branch for zero-shot semantic segmentation. Gu \etal \cite{cagnet} and Li \etal  \cite{csrl} also adopt the idea of generating synthetic features. Gu \etal \cite{cagnet} used a context-aware prior to generate features of unseen classes. Li \etal \cite{csrl} added constraints for the generation of unseen class features by exploiting structural relationships between seen and unseen categories. 

% \noindent\textbf{Spatial Information and Positional Encoding}

\section{Proposed Framework}

\subsection{Problem Formulation}
In zero-shot learning problems, class labels consist of two parts: seen categories $\mathcal{C}^S$ and unseen categories $\mathcal{C}^U$. Meanwhile, in zero-shot semantic segmentation problems, the training set $\mathcal{D}^S$ is composed of images and labels of seen categories to which image pixels belong. In other words, $\mathcal{D}^S = \{(x, y)|\ \forall i \ y_{i} \in \mathcal{C}^S \}$, where $x$ is an image and $y$ is its corresponding ground-truth label, $y_{i}$ is the ground-truth label for pixel $i$. Other images that include pixels of unseen categories are denoted by $\mathcal{D}^U = \{(x, y)|\  \exists i \  y_{i} \in \mathcal{C}^U\}$ and are only encountered at the inference time. 

\subsection{\modelfullname}
\cref{fig.network_architecture} illustrates the proposed \modelfullname (\modelname). 
\modelname\, is composed of one unlearnable mapping network \netname{M} and five learnable networks --- feature encoder (\netname{E}), generator (\netname{G}), classifier (\netname{C}), discriminator (\netname{D}), and Spatial Information Module (\netname{SIM}). During training, an input image first goes through the feature encoder \netname{E}. Then, \netname{SIM} conducts RPE on image features and produces $F_x$ for classifier \netname{C}, and a stochastic vector $z$ for generator \netname{G}. \netname{G} synthesizes features from semantic word embeddings and $z$ with a target of image features. To generalize the model on unseen categories, a random label including unseen classes is passed to \netname{M} and \netname{G}, and the synthesized features is used to train \netname{C}. At inference time, the test image only passes \netname{E}, \netname{SIM} and \netname{C}. 

The model is optimized in three stages --- (1) \textit{Training Step (SIM)} updates the main feature  encoder \netname{E}, the Spatial Information Module \netname{SIM}, and the classifier \netname{C} in a standard semantic segmentation fashion. (2) \textit{Training Step (G)} trains a generator \netname{G} that produces synthetic features with a target of real image features. (3) \textit{Transfer Learning Step} uses synthesized features to fine-tune classifier \netname{C}, which enables \netname{C} to recognize unseen categories. 

The Training Step (SIM) objective is to fine-tune the upstream feature encoder \netname{E} and train \netname{\sim}. \netname{E} is a semantic segmentation backbone network, which extracts image features. The choice of the backbone is architecture-agnostic, and any CNN-based network can be used (\eg, DeepLab \cite{chen2017deeplab}, UNet \cite{ronneberger2015u}, FCN \cite{long2015fully}). The  \simfullname(\netname{SIM}) takes image feature inputs $\netname{E}(x)$, and (1) incorporates spatial information into image features through positional encoding and (2) produces a space-aware stochastic latent representation $z$, as shown in \cref{eq.sim}, where $\oplus$ denotes concatenation operation and $PE$ stands for positional encoding vector. 
\begin{align}
    F_x,\ z &= \netname{SIM}(\netname{E}(x) \oplus PE)\label{eq.sim} 
\end{align}

We use KL divergence to force $z$ to converge to a normal distribution\cite{cagnet} to ensure stochasticity, as shown in \cref{eq.z_normal}.
\begin{align}
    \mathcal{L}_{\textit{KLD}} &= \mathrm{KL}(z \| \mathcal{N}(0,1)) \label{eq.z_normal}
\end{align}

We use the standard categorical cross-entropy loss to train classifier \netname{C}, as shown in \cref{eq:classifier_loss}
\begin{align}
    \mathcal{L}^{train}_{pred}(p, y) &= -\sum_{c} y_{c} \mathrm{log}(p_{c}) \label{eq:classifier_loss}
\end{align}
where $p = \netname{C}(F_x)$ is the categorical probabilities of features, $p_c$ is the probability of class $c$, and $y$ is the ground truth label. In Training Step (SIM), the classifier \netname{C} is trained only on real features (\ie $c\in C^S$), and the total optimization target is the weighted sum of prediction loss and KL loss for $z$, where $\alpha$ is a hyperparameter to balance losses, as shown in \cref{eq.train_sim}.
\begin{align}
    \netname{E}^*,\ \netname{\sim}^*,\ \netname{C}^* = \underset{\netname{E},\netname{\sim},\netname{C}}{\mathrm{min}} \; \mathcal{L}_{pred}^{train} + \alpha \mathcal{L}_{\textit{KLD}} \label{eq.train_sim}
\end{align}

Training Step (G) attempts to train a generator \netname{G}, with a fixed encoder \netname{E} and \netname{\sim}. The generator is needed to synthesize image features of unseen categories so that the classifier \netname{C} can recognize unseen categories after being trained on synthetic features. \netname{G} generates synthetic features from a  latent code. The latent code consists of two parts: (1) semantic word embedding $e$ and (2) normally distributed prior $z$. The stochasticity  of $z$ prevents the  generative model from collapsing, as discussed in ~\cite{zhu2017unpaired,goodfellow2014generative}. The mapping network \netname{M} maps ground truth annotations to semantic word embeddings, $e = \netname{M}(y)$. Its weights are initialized with word2vec \cite{word2vec} and fasttext \cite{fasttext}. The generator \netname{G} produces a synthetic feature $\hat{F}_x = \netname{G}(e \oplus z)$.

The synthesized features $\hat{F}_x$ have to be close to real features of seen categories. We follow previous work \cite{zs3,cagnet} and use Maximum Mean Discrepancy (MMD) loss \cite{li2015generative} to reduce the distribution distance between real and synthetic features. Total loss $\mathcal{L}_{\textit{MMD}}$ is the summation of MMD loss on seen classes $\mathcal{L}_{\textit{MMD}}(c)$, as shown in \cref{eq:mmd}. 
\begin{align}
    \mathcal{L}_{\textit{MMD}} &= \sum_c \mathcal{L}_{\textit{MMD}}(c) \quad ; \quad c \in C^S \label{eq:mmd}\\
    &\textrm{where, } \nonumber \\
    \mathcal{L}_{\textit{MMD}}(c) &= \sum_{f,f'\in F_{x,c}} k(f, f') + \sum_{\hat{f},\hat{f'}\in \hat{F}_{x,c}} k(\hat{f}, \hat{f'}) \nonumber \\ 
    &~~~~~~~~~~~~- 2\sum_{f \in F_{x,c}} \sum_{\hat{f}\in \hat{F}_{x,c}} k(f, \hat{f})
\end{align}
where $F_{x,c}$ and $\hat{F}_{x,c}$ are real and synthetic features for class $c$ in sample $x$'s feature, respectively. We choose Gaussian kernel function $k(f,f')=\mathrm{exp}(-\frac{1}{2} \| f - f'\|^2)$ as suggested in \cite{zs3}. In order to make the synthesized image features realistic, we add a discriminator \netname{D} and training \netname{G} to deceive \netname{D} by optimizing an adversarial loss, as shown in~\cref{eq:adersarial_loss}~ \cite{goodfellow2014generative}.
\begin{align}
    \mathcal{L}_{adv} = \mathbb{E}_{f \in F_x}[\mathrm{log}(\netname{D}(f))] + \mathbb{E}_{\hat{f} \in \hat{F}_x}[\mathrm{log}(1-\netname{D}(\hat{f}))] \label{eq:adersarial_loss}
\end{align}

The total loss for Training Step (G) is composed of MMD loss and adversarial loss, and hyperparameter $\beta$ controls the trade-off between two losses, as shown in \cref{eq.train_G}.
\begin{align}
    \netname{G}^*,\ \netname{D}^* = \underset{\netname{G}}{\mathrm{min}} \, \underset{\netname{D}}{\mathrm{max}} \; \mathcal{L}_{adv} + \beta \mathcal{L}_{\textit{MMD}} \label{eq.train_G}
\end{align}

Since the trainable networks and the losses do not overlap in \cref{eq.train_sim,eq.train_G}, we jointly optimize them for  efficiency. Finally, to synthesize features for unseen categories, during the Transfer Learning Step, a pseudo-ground-truth $\hat{y}$ (\ie, a pseudo label including unseen categories) is fed into \netname{M} and \netname{G}. The synthetic features are used to train the classifier \netname{C} so that \netname{C} can recognize the unseen categories. The prediction loss of the Transfer Learning Step is shown in \cref{eq.trans_pred_loss}. 
\begin{align}
    \mathcal{L}_{pred}^{trans}(p, \hat{y}) &= -\sum_{c} \widehat{y_{c}} \mathrm{log}(p_{c}) \label{eq.trans_pred_loss}
\end{align}

The classifier \netname{C} is optimized on synthetic features as well as real features to avoid performance drop on seen categories, as shown in \cref{eq.train_transfer}. 
\begin{align}
    \netname{C}^* = \underset{\netname{C}}{\mathrm{min}} &\; \mathcal{L}_{pred}^{train} + \mathcal{L}_{pred}^{trans} \label{eq.train_transfer}
\end{align}

\subsection{Relative Positional Encoding}
As illustrated in \cref{fig.network_architecture}(d), the Spatial Information Module consists of the encoding module and the reparameterization module. The encoding module uses a residual structure \cite{he2016deep} and incorporates spatial information into image features using \textit{positional encoding}~\cite{vaswani2017attention}. The reparameterization module~\cite{kingma2013auto} takes the output of the encoding module and generates a stochastic latent code. In \cref{sec.sim_architecture}, we discuss and compare different architectures for \sim.

RPE uses sine and cosine functions to incorporate pixel positions into a feature map \cite{vaswani2017attention}. To handle 2D positional encoding, we use a 600-dimensional vector, in which the first 300 dimensions are used for horizontal location encoding and the last 300 dimensions are used for vertical location encoding. \cref{eq.positional_encoding_horizontal1,eq.positional_encoding_horizontal2} show horizontal positional encoding and \cref{eq.positional_encoding_vertical1,eq.positional_encoding_vertical2} show vertical positional encoding. $^{i}pos_u^*$ and $^{i}pos_v^*$ represent the relative horizontal and vertical positions of pixel $i$, respectively, and $d$ denotes the dimension. The overall dimensionality of positional encoding in each direction is  $d_{model} = 300$.
\begin{align}
    PE(^{i}pos_u^{*}, 2d) &= \mathrm{sin}(^{i}pos_u^{*}/10000^{2d/d_{model}})\label{eq.positional_encoding_horizontal1} \\
    PE(^{i}pos_u^{*}, 2d+1) &= \mathrm{cos}(^{i}pos_u^{*}/10000^{2d/d_{model}})\label{eq.positional_encoding_horizontal2} \\
    PE(^{i}pos_v^{*}, 2d) &= \mathrm{sin}(^{i}pos_v^{*}/10000^{2d/d_{model}})\label{eq.positional_encoding_vertical1} \\
    PE(^{i}pos_v^{*}, 2d+1) &= \mathrm{cos}(^{i}pos_v^{*}/10000^{2d/d_{model}})\label{eq.positional_encoding_vertical2}
\end{align}

In order to handle arbitrary image sizes, we do not use the absolute position $pos$ of a given pixel in the feature map. Rather, we use the relative position $pos^*$, as shown in \cref{eq.relative_pe1} and \cref{eq.relative_pe2}.
\begin{align}
    ^{i}pos_u^{*} &= c \cdot\, ^{i}pos_u / W\label{eq.relative_pe1} \\
    ^{i}pos_v^{*} &= c \cdot\, ^{i}pos_v / H\label{eq.relative_pe2}
\end{align}
where $c$ is a constant which we set to $512$, and $H$ and $W$ are the height and width of the image feature, respectively. Please note that despite using the same name, our RPE is different from the ones ~\cite{dai2019transformer,huang2020improve} used in natural language problems, which use pair-wise token relation for positional encoding and is fundamentally different from positional encoding for image features. 

\subsection{Annealed Self-Training}
In prior literature \cite{zhu2005semi}, self-training was used to leverage the model's prediction on unlabeled samples to obtain additional \emph{pseudo-annotations} for fine-tuning the model. In zero-shot segmentation, the model produces class labels and confidence values (\ie, output of softmax layer) upon encountering pixels of unseen classes. The produced class labels are used as pseudo labels for self-training to learn unseen classes, based on the output confidence values. We anticipate that the generated pseudo-labeled pixels may be incorrect, and therefore, noisy labels may degrade the performance of the segmentation model. Previous methods~\cite{zs3,cagnet} threshold the prediction confidence and use only high confidence pseudo labels (\eg, highest 75\% in \cite{zs3}) during training to reduce the influence of incorrect pseudo-annotations.

However, finding a suitable threshold is not trivial since the model's confidence in each sample is different. Inspired by knowledge distillation in transfer learning \cite{hinton2015distilling}, we propose Annealed Self-Training (AST), which uses all pseudo-annotations but assigns different loss weights according to the confidence score, as shown in \cref{eq:ast}
\begin{align}
    \vspace{-10px} 
    w_i = {1 \over Z} {\textrm{exp}(p_i/T) \over \sum_i \textrm{exp}(p_i/T)} \label{eq:ast}
    \vspace{-10px}
\end{align}
where $Z$ is a normalization term so that the maximum value of loss weights ($\textrm{max}\{w_i\}$) is 1. The loss re-weighing is achieved by applying \textit{annealing softmax} function on confidence score $p$, and the annealed temperature $T$ is used to adjust re-weighing intensity. Note that we only do loss re-weighing on pseudo-annotations and the loss weights of seen classes are always 1. 

\section{Experimental Evaluation}

\begin{table*}%[!h]
    % \vspace{-5px}
    \centering
    \small
    \caption{Zero-shot semantic segmentation mIoU performance on Pascal VOC, Pascal Context and COCO Stuff. ``ST'' and ``AST'' stand for self-training and annealed self-training, respectively. Evaluation metric is mean intersection over union (mIoU)}
    \label{tab.quantitative_result}
    \setlength{\tabcolsep}{0.3em} % for horizontal padding
    \begin{tabular}{l|ccc|ccc|ccc}
         \toprule
         & \multicolumn{3}{|c}{Pascal VOC} & \multicolumn{3}{|c}{Pascal Context} & \multicolumn{3}{|c}{COCO Stuff} \\
         \midrule
         \bf{Methods} & \bf{Seen(\%)} & \bf{Unseen(\%)} & \bf{Harmonic(\%)} & \bf{Seen(\%)} & \bf{Unseen(\%)} & \bf{Harmonic(\%)} & \bf{Seen(\%)} & \bf{Unseen(\%)} & \bf{Harmonic(\%)}\\
         \cmidrule(r){1-1} \cmidrule(lr){2-2} \cmidrule(lr){3-3} \cmidrule(lr){4-4} \cmidrule(lr){5-5} \cmidrule(lr){6-6} \cmidrule(lr){7-7} \cmidrule(lr){8-8} \cmidrule(lr){9-9} \cmidrule(l){10-10}
         SPNet~\cite{spnet} & 78.00 & 15.63 & 26.10 & 35.14 & 4.00 & 7.18 & \textbf{35.18} & 8.73 & 13.98 \\
         ZS3~\cite{zs3} &  77.30 & 17.65 & 28.74 & 33.04 & 7.68 & 12.46 & 34.66 & 9.53 & 14.95\\
         CaGNet~\cite{cagnet} & \textbf{78.40} & 26.59 & 39.72 & \textbf{36.10} & 14.42 & 20.61 & 33.49 & 12.23 & 18.19 \\
         \cmidrule(r){1-1} \cmidrule(lr){2-2} \cmidrule(lr){3-3} \cmidrule(lr){4-4} \cmidrule(lr){5-5} \cmidrule(lr){6-6} \cmidrule(lr){7-7} \cmidrule(lr){8-8} \cmidrule(lr){9-9} \cmidrule(l){10-10}
         SIGN (Ours) & 75.40 & \textbf{28.86} & \textbf{41.74} & 33.67 & \textbf{14.93} & \textbf{20.67} & 32.31 & \textbf{15.47} & \textbf{20.93} \\
         \midrule\midrule
         ZS3 + ST & 78.02 & 21.15 & 33.28 & 33.98 & 9.53 & 14.88 & 34.89 & 10.55 & 16.20 \\
         CaGNet + ST & 78.59 & 30.31 & 43.66 & \textbf{36.44} & 16.30 & 22.52 & 35.55 & 13.40 & 19.46 \\
         \cmidrule(r){1-1} \cmidrule(lr){2-2} \cmidrule(lr){3-3} \cmidrule(lr){4-4} \cmidrule(lr){5-5} \cmidrule(lr){6-6} \cmidrule(lr){7-7} \cmidrule(lr){8-8} \cmidrule(lr){9-9} \cmidrule(l){10-10}
         SIGN + ST & 78.62 & 33.12 & 46.61 & 34.91 & 16.71 & 22.60 & \textbf{36.39} & 15.15 & 21.39\\
         SIGN + AST & \textbf{83.49} & \textbf{41.29} & \textbf{55.26} & 34.90 & \textbf{17.86} & \textbf{23.62} & 31.94 & \textbf{17.53} & \textbf{22.64} \\
         \bottomrule
    \end{tabular}

\end{table*}

\begin{figure*}
    % \vspace{-5px}
    \centering
    \includegraphics[width=0.98\linewidth]{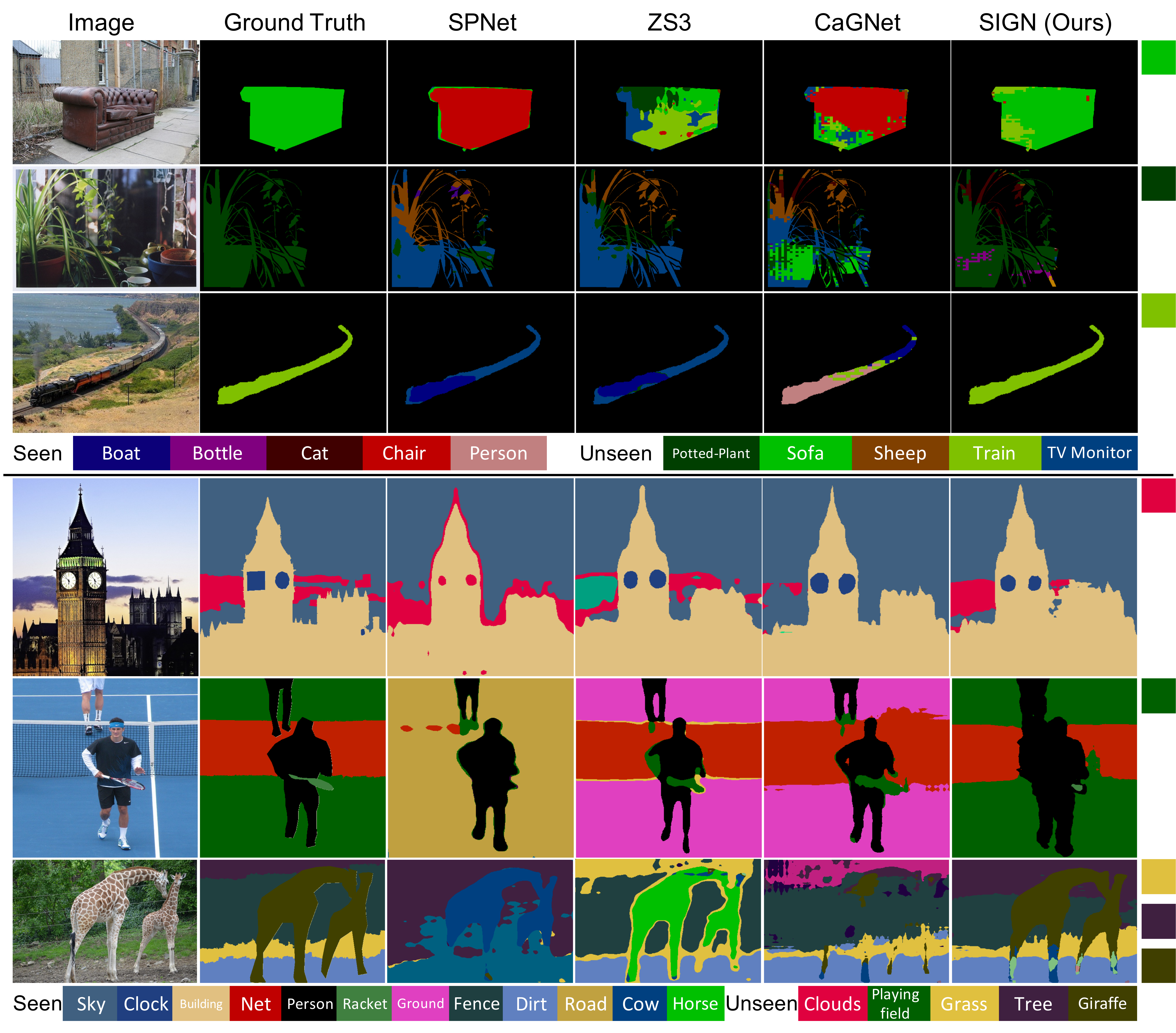}
    \caption{Qualitative comparison with SPNet \cite{spnet}, ZS3 \cite{zs3} and CaGNet \cite{cagnet}. The top three samples are from Pascal VOC and the bottom three samples are from COCO Stuff. Color bar below the samples indicate the correspondence between colors and categories (including false positive categories). The square(s) on the right indicates the unseen class(es) in the sample on the left. }
    \label{fig.qualitative_baseline}
    % \vspace{-10px}
\end{figure*}

\subsection{Benchmark Datasets}

Following \cite{spnet,cagnet}, we used (1) Pascal Visual Object Classes (VOC)  \cite{everingham2010pascal}, (2) Pascal Context \cite{mottaghi2014role} and (3) COCO Stuff \cite{caesar2018coco} for evaluation. Pascal VOC contains 20 categories with 1,464 and 1,449 images, for training and testing, respectively. Since Pascal VOC is relatively small, external Semantic Boundaries Dataset (SBD) dataset \cite{BharathICCV2011} is also used during training as suggested in previous works \cite{spnet,zs3,cagnet}. After introducing SBD and excluding duplicate images in Pascal VOC test set, there are 8,284 and 2,299 images for training and validation, respectively. Pascal Context contains 33 categories, including 4998, 500 and 5105 images, for training, validation and testing. COCO Stuff is a large semantic segmentation dataset with 171 categories. There are 118,287 images for training and 5,000 for testing. We split the last 10,000 images in the training set for validation.

We follow the evaluation protocol from \cite{spnet,cagnet} for splitting seen and unseen categories. For Pascal VOC, the last five classes (potted plant, sheep, sofa, train, tv-monitor) are used as unseen categories \cite{spnet,cagnet}. Class ``background'' is ignored in Pascal VOC during both training and testing, as suggested by \cite{cagnet}, since it is unreasonable to use single semantic word representation for all kinds of background objects (\eg, sky, road). Four categories (cow, motorbike, sofa, cat) are classified as unseen classes in Pascal Context \cite{cagnet}. For COCO Stuff, 15 classes (frisbee, skateboard, cardboard, carrot, scissors, suitcase, giraffe, cow, road, wallconcrete, tree, grass, river, clouds, playingfield) are treated as unseen \cite{spnet,cagnet}. 

\subsection{Experiment Setup And Evaluation Metrics}

\noindent\textbf{Implementation Details:} We use word embeddings from both word2vec \cite{word2vec} and fasttext \cite{fasttext}, and concatenate them to represent words (a total of 600 dimensional vector; 300 for each). We follow \cite{cagnet} to use average word embeddings when a category has multiple words. We use Deeplab-v2 \cite{chen2017deeplab} built upon ResNet-101 \cite{he2016deep} as the semantic segmentation backbone. We use SGD \cite{rezende2014stochastic} optimizer for Deeplab-v2 backbone, \sim and classifier with initial learning rate $2.5 \times 10^{-4}$, and Adam \cite{kingma2014adam} optimizer for the generator with initial learning rate $2 \times 10^{-4}$. A poly learning rate scheduler was applied for backbone as suggested by  \cite{spnet}. We empirically set loss weights to  $\alpha=100$ and $\beta=50$.

\noindent\textbf{Evaluation Metrics:} We report performance based on mean intersection-over-union (mIoU) and conduct evaluation under generalized zero-shot learning (GZSL) metric. Generalized zero-shot learning assesses performance on seen and unseen classes at the same time, rather than  evaluating only on the unseen classes under zero-shot learning (ZSL) metric. 
Similar to previous work on zero-shot classification \cite{xian2017zero} and segmentation \cite{spnet,zs3}, we report mIoU of seen and unseen categories and harmonic mIoU of seen and unseen categories. 

\subsection{Comparison With Baselines:} We compare our method against (1) SPNet \cite{spnet}, (2) ZS3 \cite{zs3} and (3) CaGNet \cite{cagnet}. We do not compare with CSRL \cite{csrl} since they use a different protocol and their code is not available. For fair comparison, we use the same word2vec \cite{word2vec} and fasttext \cite{fasttext} embeddings, and use the same Deeplab-v2 \cite{chen2017deeplab} as the segmentation backbone for all the methods. 

We report the performance with and without self-training. \cref{tab.quantitative_result} summarizes the results of different methods. We can see that our \modelname\, model achieves the best performance on unseen categories and harmonic mIoU on all the three benchmark datasets, which indicates the effectiveness of our method on recognizing unseen categories. In addition, our Annealed Self-Training further improves performance over conventional self-training. Compared to seen categories, AST works better for unseen ones due to higher utilization of pseudo-annotations. We notice that the performance impact of AST on Pascal VOC is higher than Context and COCO Stuff. Performance difference can be attributed to the smaller number of categories in Pascal VOC, which leads to a higher chance of correct pseudo labels.

\cref{fig.qualitative_baseline} shows a qualitative comparison between \modelname\, and baselines. We can see that \modelname\, achieves high accuracy, even when there are multiple unseen categories (please see last row). 

\subsection{Ablation Studies}
We conduct ablation studies on Pascal VOC dataset to show the effectiveness of Relative Positional Encoding and Annealed Self-Training.

\noindent\textbf{\sim Architecture:}\label{sec.sim_architecture}
We designed \sim as a residual module \cite{he2016deep} and experimented with four  architectures. (A) Convolution-based \sim uses simple residual blocks  \cite{he2016deep} with consecutive convolution and activation layers. (B) Attention-based  \cite{cagnet} \sim learns three attention maps of different scales from deep features. Deep features are then element-wise multiplied with attention maps, and concatenated together. (C) Self-attention-based \sim uses the structure of Transformer Encoder \cite{vaswani2017attention}. The difference between self-attention- and attention-based \sim is: (1) attention map is computed according to the correlation of pixels on deep features instead of pixel-wise attention, and (2) feature aggregation computes the weighted sum of previous deep features, rather than concatenating features of different scales. (D) Multihead self-attention-based \cite{vaswani2017attention} runs several self-attention in parallel. The input feature is first linearly transformed into smaller dimension in each head, and self-attention is applied separately. Then, the attention results are concatenated together and linearly transformed back to the original dimension. 

\cref{tab.sim_num_of_param} summarizes the number of parameters of the four \sim architectures and their performance on Pascal VOC. The best performance is achieved by multihead self-attention-based \sim, followed by attention-based \sim. According to the performance in \cref{tab.sim_num_of_param},  performance numbers in all following experiments are reported based on multihead self-attention-based \sim. Please refer to \cref{sec.detailed_sim} for the detailed model structure. 

\begin{table}[!h]
    % \vspace{-5px}
    \centering
    \small
    \caption{Number of parameters in four different \sim architecture and the corresponding performance on Pascal VOC.}
    \label{tab.sim_num_of_param}
    \setlength{\tabcolsep}{0.3em}
    \begin{tabular}{C{1.6cm}C{1.3cm}C{1.5cm}C{1.5cm}C{1.5cm}}
         \toprule
         \bf{Archi.} & \bf{Conv} & \bf{Attention} & \bf{Self-Attn.} & \bf{Multi SA} \\
         \cmidrule(r){1-1} \cmidrule(lr){2-2} \cmidrule(lr){3-3} \cmidrule(lr){4-4} \cmidrule(l){5-5}
         \# Params & 5.25M & 5.24M & 4.60M & 4.86M \\
         \cmidrule(r){1-1} \cmidrule(lr){2-2} \cmidrule(lr){3-3} \cmidrule(lr){4-4} \cmidrule(l){5-5}
         H. mIoU(\%) & 39.13 & 40.86 & 40.51 & \textbf{41.74} \\
         \bottomrule
    \end{tabular}
    % \vspace{-10px}
\end{table}

\noindent\textbf{Relative versus Absolute Positional Encoding:}
To evaluate the effectiveness of the proposed Relative Positional Encoding, we compare it with two other positional encoding strategies: (1) Absolute Positional Encoding (APE), which use the absolute index of pixel to compute positional vector, and (2) Absolute Positional Encoding with Interpolation during testing \cite{dosovitskiy2020image}, which computes positional vector based on the training image size and does bilinear interpolation on positional vector during testing. 

The results are presented in \cref{tab.relative_vs_absolute}. We see that on unseen categories mIoU and harmonic mIoU, RPE improves performance by 3\% compared to APE. Adding bilinear interpolation to APE improves performance by roughly 2\% but still cannot match the performance of RPE. Interestingly, we notice a performance degradation of APE compared to the model without PE. We speculate that this is due to the mismatch between training and test image size, and due to the larger test image size, APE fails to encode all spatial information. Please note that larger image sizes or even multi-scale input sizes are commonly used during testing, because it can provide better prediction performance \cite{long2015fully,chen2017deeplab}. 

\begin{table}[]
    \vspace{-5px}
    \centering
    \small
    \caption{Mean IoU of model without PE, Absolute PE, Absolute PE with interpolation and Relative PE on Pascal VOC. Numbers in parentheses show the improvement over model without PE.}
    \label{tab.relative_vs_absolute}
    \setlength{\tabcolsep}{0.3em}
    \begin{tabular}{lccc}
         \toprule

         \bf{Methods} & \bf{Seen(\%)} & \bf{Unseen(\%)} & \bf{Harmonic(\%)} \\
         \cmidrule(r){1-1} \cmidrule(lr){2-2} \cmidrule(lr){3-3} \cmidrule(l){4-4}
         w/o PE & 71.86 & 26.07 & 38.26 \\
         APE & 70.44 & 25.68 & 37.64 \\
         APE w/ Inter. & 71.17 & 27.36 & 39.53 \\
         RPE & \textbf{75.40} (+3.54) & \textbf{28.86} (+2.79) & \textbf{41.74} (+3.21) \\
         \bottomrule
    \end{tabular}
    \vspace{-10px}
\end{table}

\begin{figure}
    \centering
    \includegraphics[width=\linewidth]{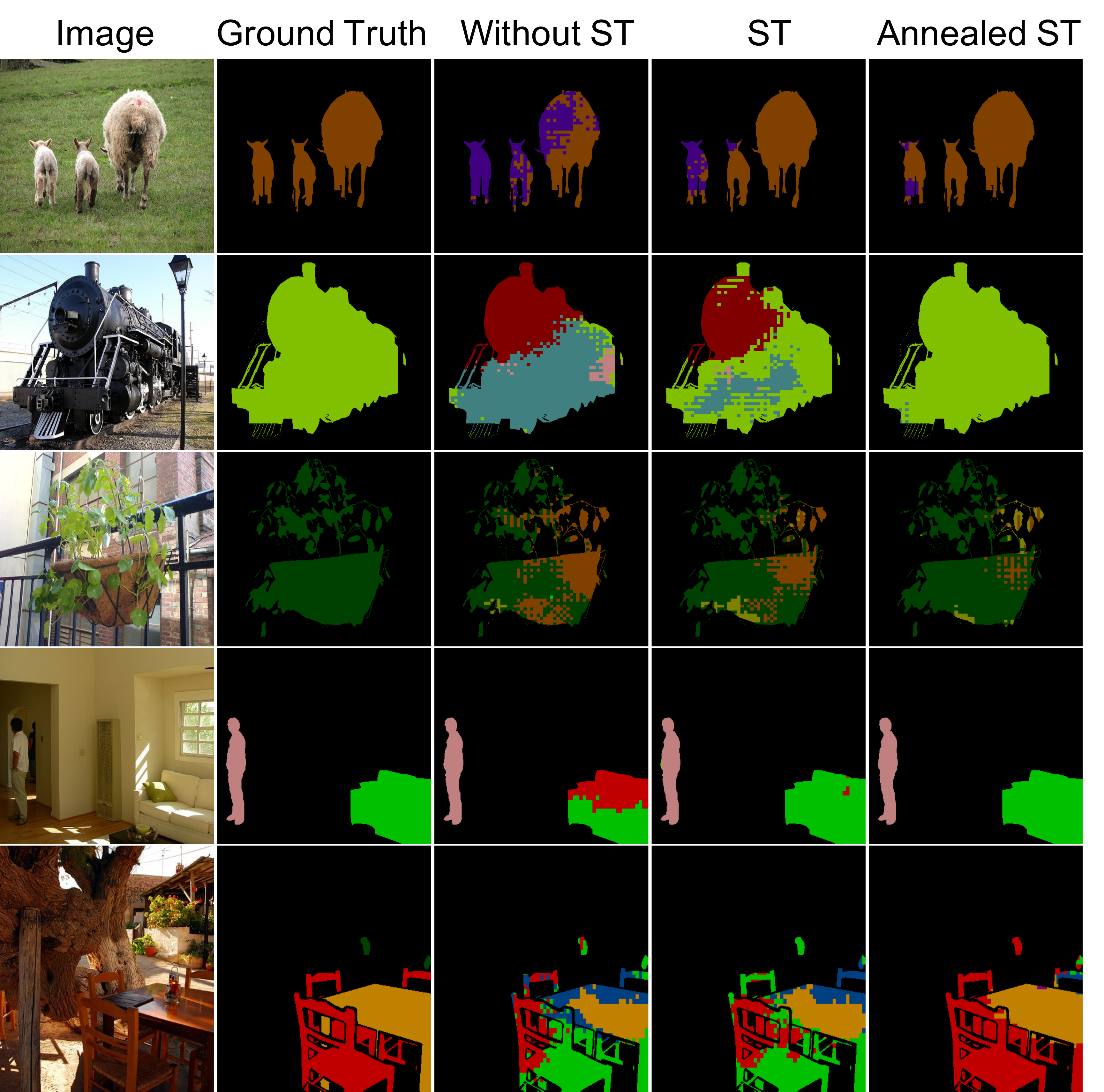}
    \caption{Qualitative comparison of predictions without self-training, with self-training in \cite{zs3} and our Annealed Self-Training}
    \label{fig.qualitative_ast}
    \vspace{-10px}
\end{figure}

\begin{figure}
    \vspace{-5px}  
    \centering
    \includegraphics[width=0.8\linewidth]{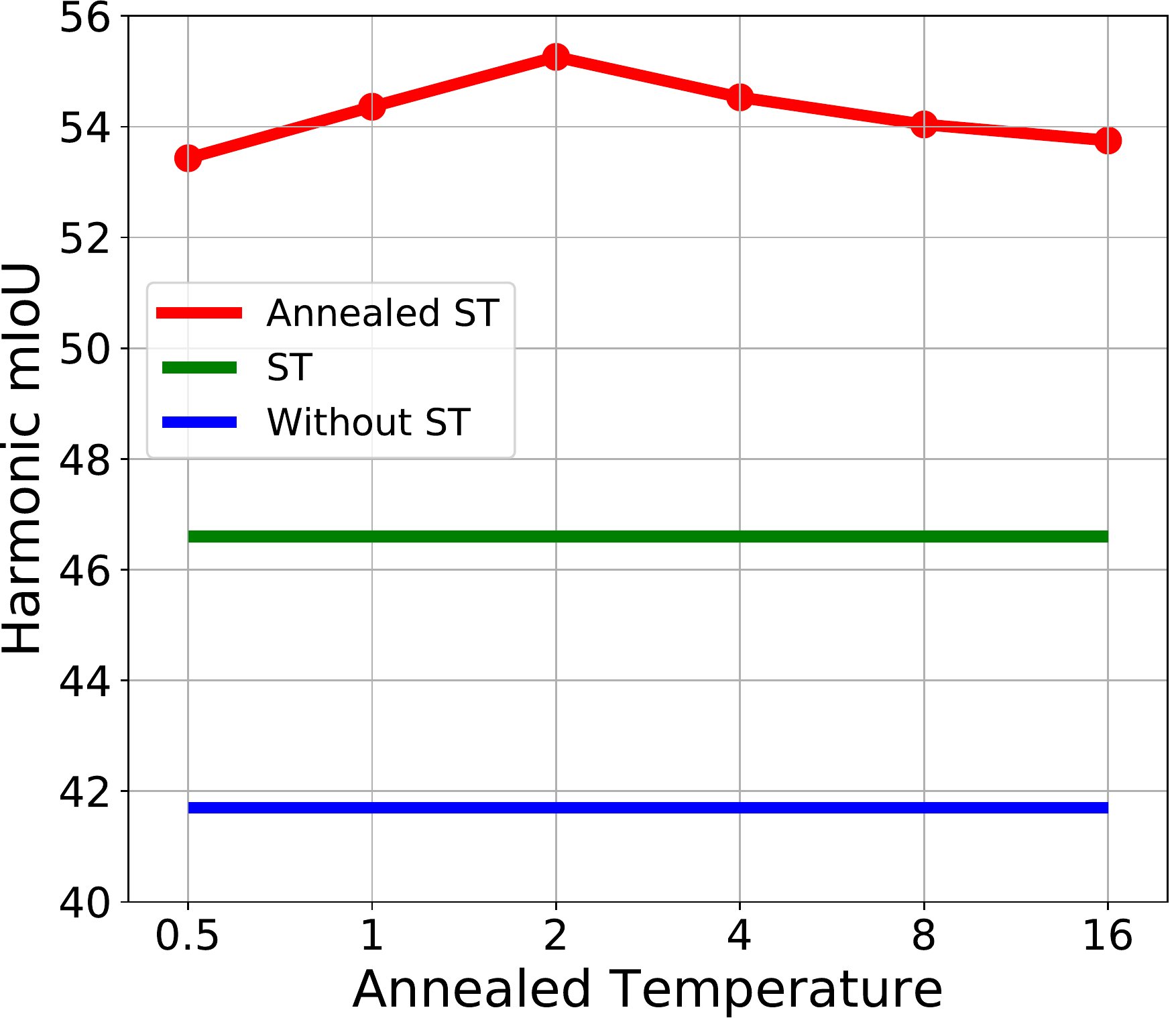}
    \caption{Harmonic mIoU on Pascal VOC under different annealed temperature in Annealed Self-Training (red line).}
    \label{fig.annealed_st}
    \vspace{-10px}
\end{figure}

\noindent\textbf{Effect Of Annealed Temperature:}
In annealed self-training, pseudo-annotations with higher confidence are assigned higher weights in the loss calculation. In the \textit{softmax} function, the annealed temperature controls the smoothness of the output. The higher the temperature, the smoother the output. We adjust the loss weights assigned to high-confidence and low-confidence pseudo-annotations by changing the annealed temperature. \cref{fig.annealed_st} shows the performance curve \wrt to annealed temperature on Pascal VOC. We noticed that a temperature of 2 is empirically optimal, and different temperature values leads to lower performance. The reasons for this observation could include (1) for low temperatures, \textit{softmax} function produces sharp output which completely ignores the pseudo-annotations with low confidence, and (2) for high temperatures, due to the smoothness of the output of the \textit{softmax} function, the loss weights assigned to high-confidence and low-confidence pseudo-annotations are too close.

\noindent\textbf{Impact Of Spatial Information On Semantic Segmentation:}
%In this study, we show that adding spatial information is also beneficial to conventional semantic segmentation problems. 
We tested all four structures of \sim on seen categories before the transfer learning step. In the Deeplab-v2 model without \sim, we slightly increased the number of trainable parameters to ensure that the total parameters are roughly the same as models with \sim. \cref{tab.spatial_info_benifit} shows that even with the simplest convolutional \sim (second column), the performance is improved after adding spatial information. For the \sim based on self-attention and multi-head self-attention, greater improvements can be seen (last two columns). Our hypothesis for this observation is that the self-attention mechanism relaxes the restrictions on the receiving field, so it can make more effective use of spatial information.

\begin{table}[!h]
    \centering
    \small
    \caption{mIoU(\%) on seen categories before transfer learning on Pascal VOC.}
    \label{tab.spatial_info_benifit}
    \setlength{\tabcolsep}{0.3em}
    \begin{tabular}{C{1.2cm}C{1.2cm}C{1.2cm}C{1.2cm}C{1.4cm}}
         \toprule
         \bf{w/o \sim} & \bf{Conv} & \bf{Attention} & \bf{Self-att.} & \bf{Multi SA} \\
         \cmidrule(r){1-1} \cmidrule(lr){2-2} \cmidrule(lr){3-3} \cmidrule(lr){4-4} \cmidrule(l){5-5} 
        76.59 & 76.83 & 76.97 & 77.75 & \textbf{77.87} \\
        %  Context & 34.39 & x & x & x & x \\
        %  COCO & 33.57 & x & x & x & x \\
         \bottomrule
    \end{tabular}
    % \vspace{-10px}
\end{table}

\vspace{-5px}
\section{Conclusion}
\vspace{-5px}
We proposed a new zero-shot semantic segmentation framework that incorporates spatial information into prediction. Our method is flexible to handle varying image size using a novel Relative Positional Encoding scheme. We introduced a new self training strategy - Annealed Self Training, which automatically adjusts the importance of pseudo-annotations from prediction confidence. We conducted an extensive experimental study and validated the effectiveness of the proposed RPE and AST, and also investigated network architectures for encoding spatial information. Finally, our \modelname\, model showed state-of-the-art performance for zero-shot semantic segmentation on benchmark datasets and has the potential to improve performance on conventional semantic segmentation problems. 

\noindent\small\textbf{Acknowledgement} This material is based on research sponsored by Air Force Research Laboratory (AFRL) under agreement number FA8750-19-1-1000. The U.S.Government is authorized to reproduce and distribute reprints for Government purposes notwithstanding any copyright notation therein. The views and conclusions contained herein are those of the authors and should not be interpreted as necessarily representing the official policies or endorsements, eigher expressed or implied, of Air Force Laboratory, DARPA or the U.S. Government.

{\small
\bibliographystyle{ieee_fullname}
\bibliography{egbib}
}
\clearpage

\appendix

\twocolumn[{%
\renewcommand\twocolumn[1][]{#1}%
\begin{center}
\textbf{\Large SIGN: Spatial-information Incorporated Generative Network \\ for Generalized Zero-shot Semantic Segmentation}

\vspace{10pt}
\textbf{\large Supplementary Material}
\vspace{10pt}
\end{center}%
}]

\section{Detailed SIM Structure}\label{sec.detailed_sim}
\begin{figure}[!h]
    \centering
    \includegraphics[width=\linewidth]{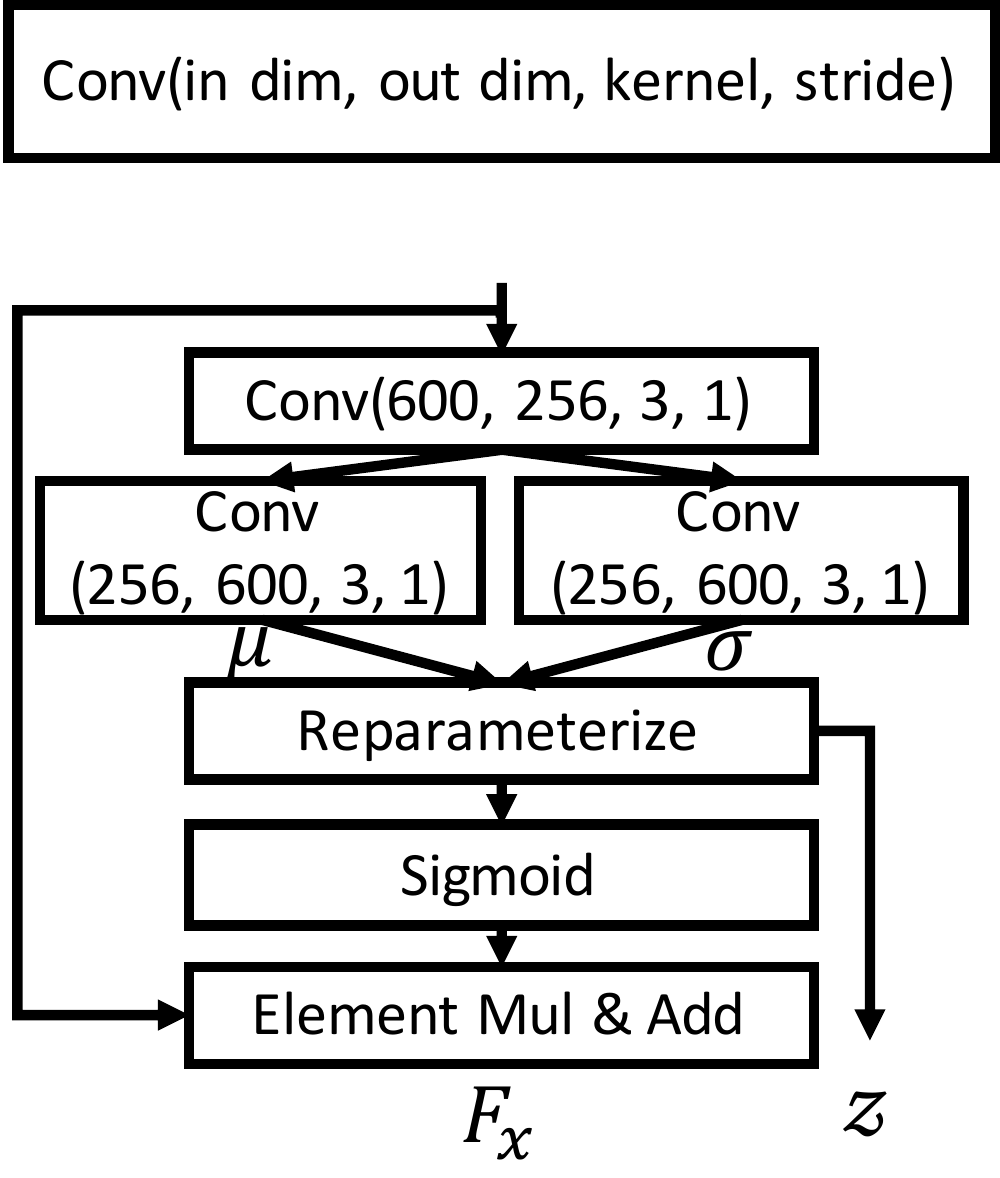}
    \caption{Reparameterization module used in \sim. }
    \label{fig.post_process_module}
\end{figure}

\begin{figure}[!h]
    \centering
    \includegraphics[width=\linewidth]{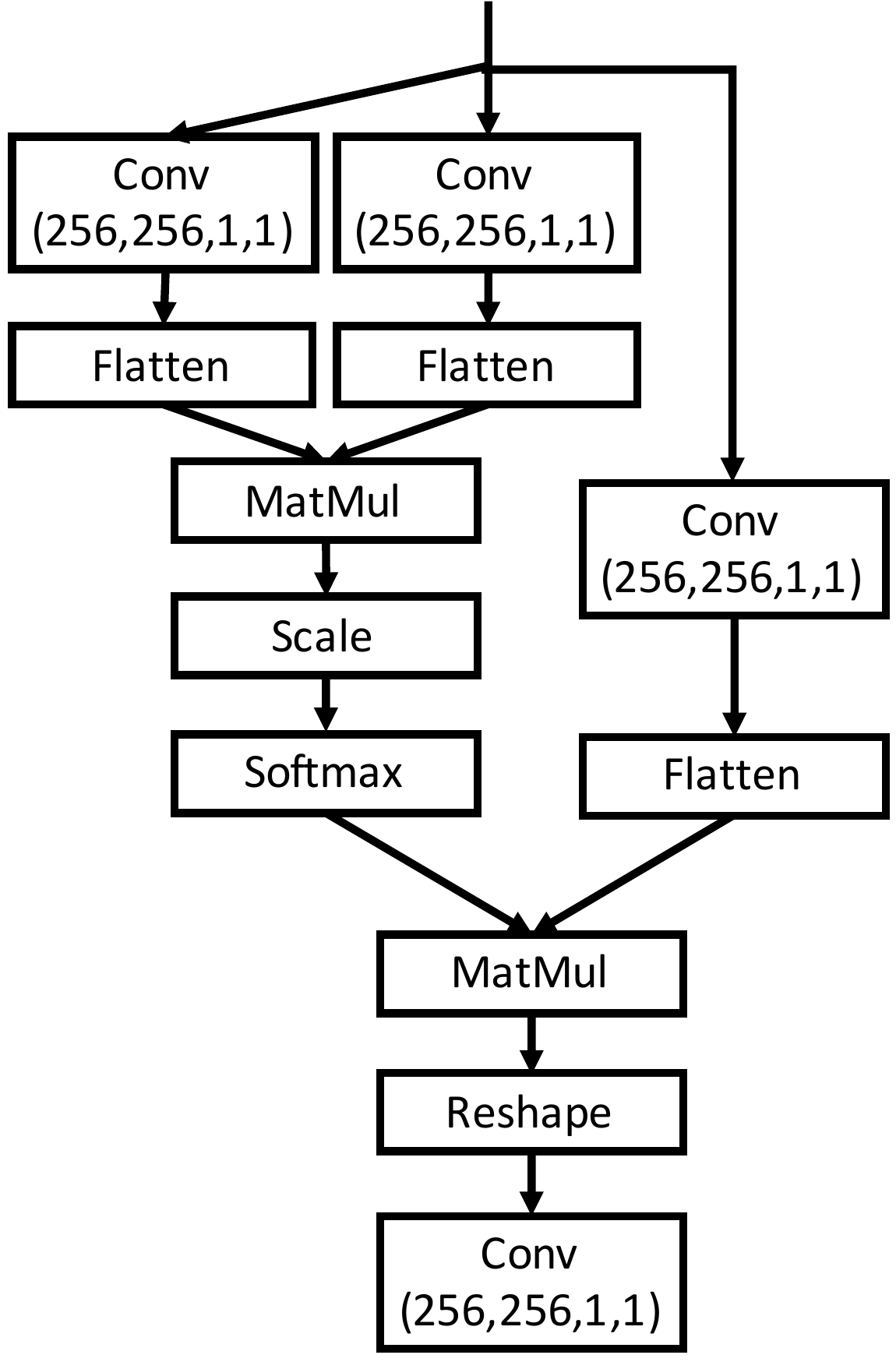}
    \caption{Self-attention module used in \sim}
    \label{fig.self_attn_module}
\end{figure}

\begin{figure*}[!h]
    \centering
    \includegraphics[width=\linewidth]{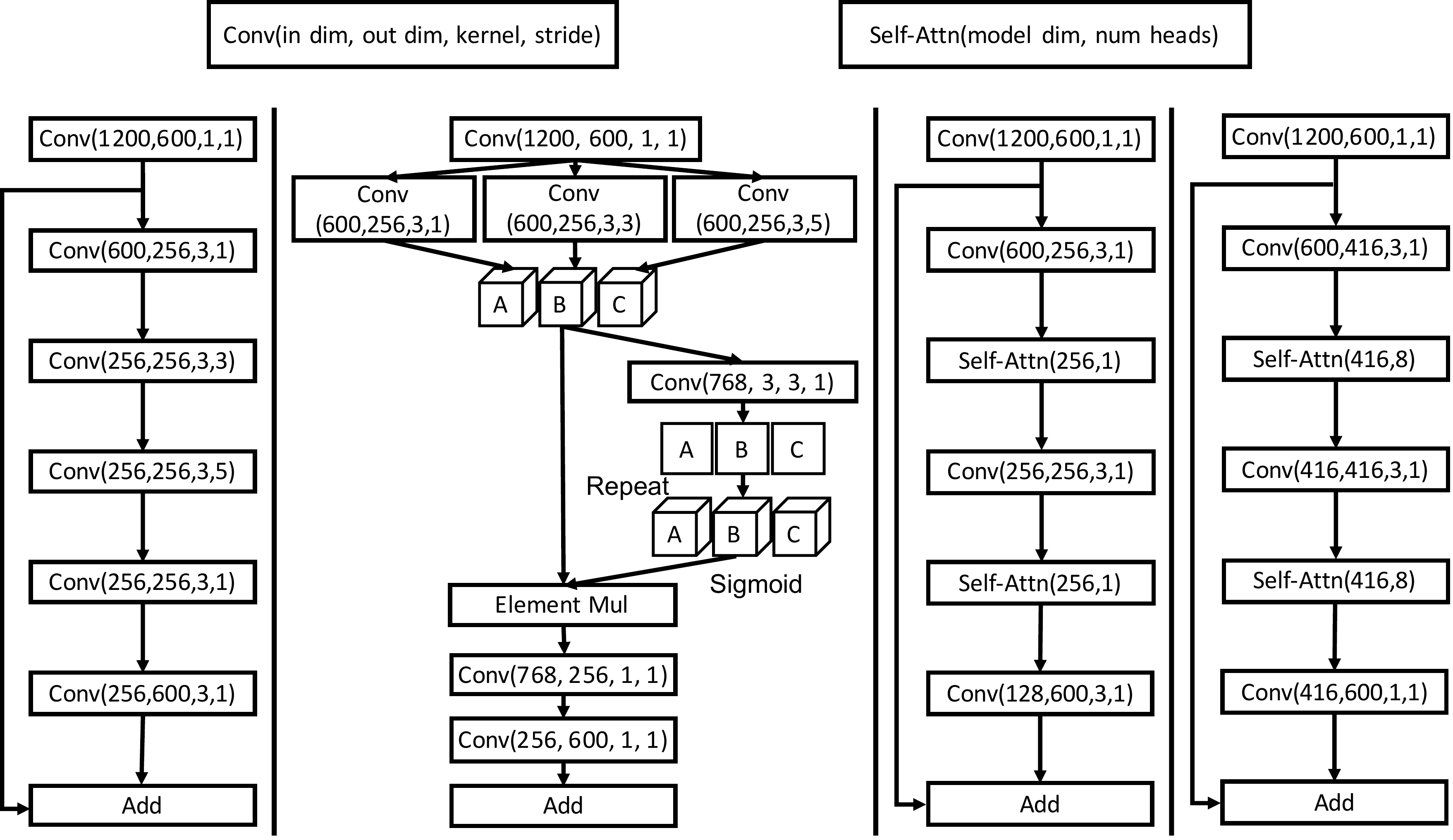}
    \caption{Four architectures tested in \sim. a) Convolution-based\cite{he2016deep}. b) Attention-based\cite{cagnet}. c) Self-attention-based\cite{vaswani2017attention}. d) Multihead self-attention-based\cite{vaswani2017attention}}
    \label{fig.arch_spatial_module}
\end{figure*}

\cref{fig.post_process_module} shows the reparameterization module in Spatial Information Module. \cref{fig.self_attn_module} shows the self-attention module and \cref{fig.arch_spatial_module} shows the details of four architectures tested for \sim in \cref{sec.sim_architecture}.

\section{Additional Qualitative Results}
\begin{figure*}
    \centering
    \includegraphics[width=0.8\linewidth]{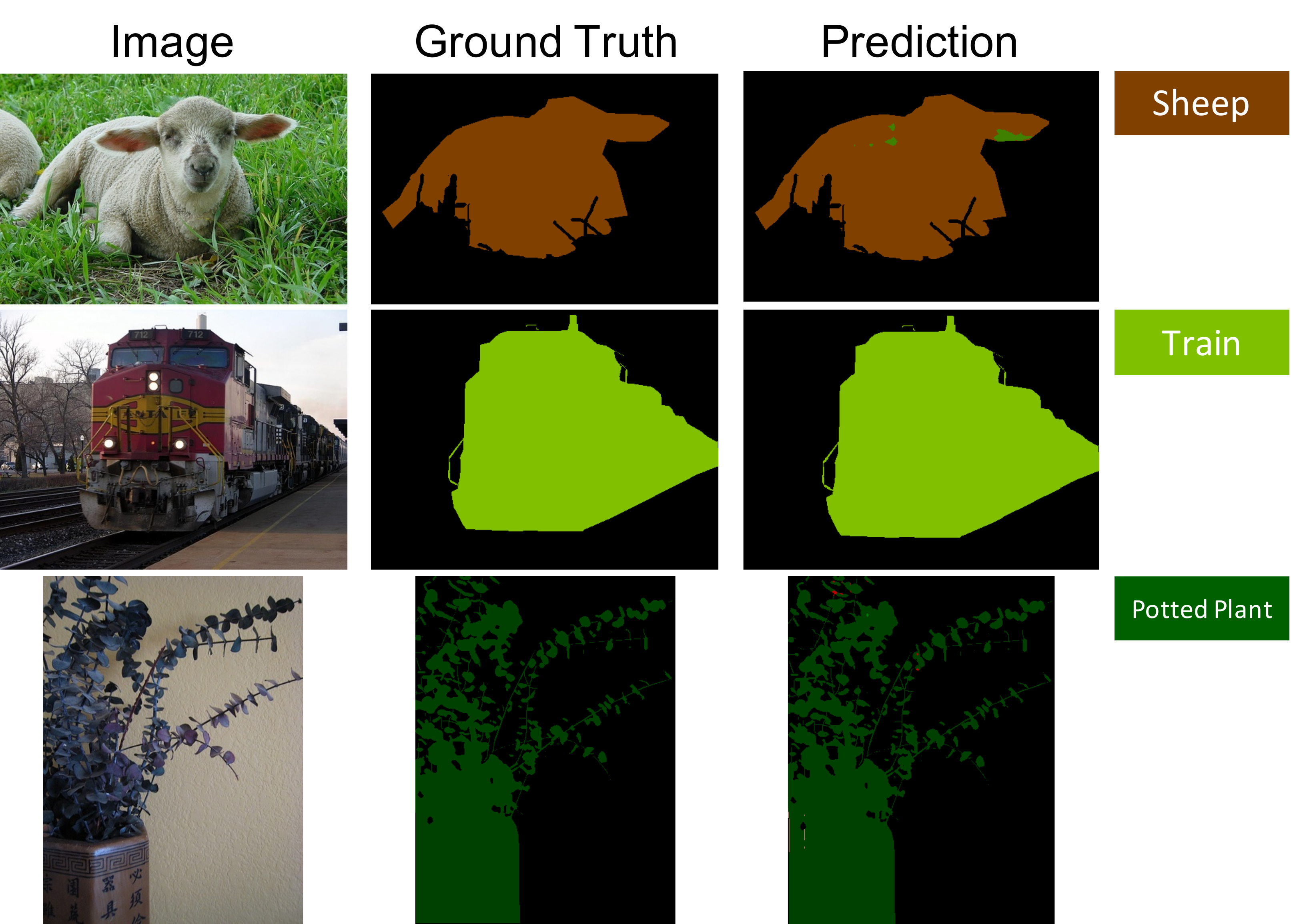}
    \caption{Additional qualitative results on Pascal VOC. Rectangle on the right side indicates the unseen class in the sample on the left}
    \label{fig.voc_supp}
\end{figure*}

\begin{figure*}
    \centering
    \includegraphics[width=0.8\linewidth]{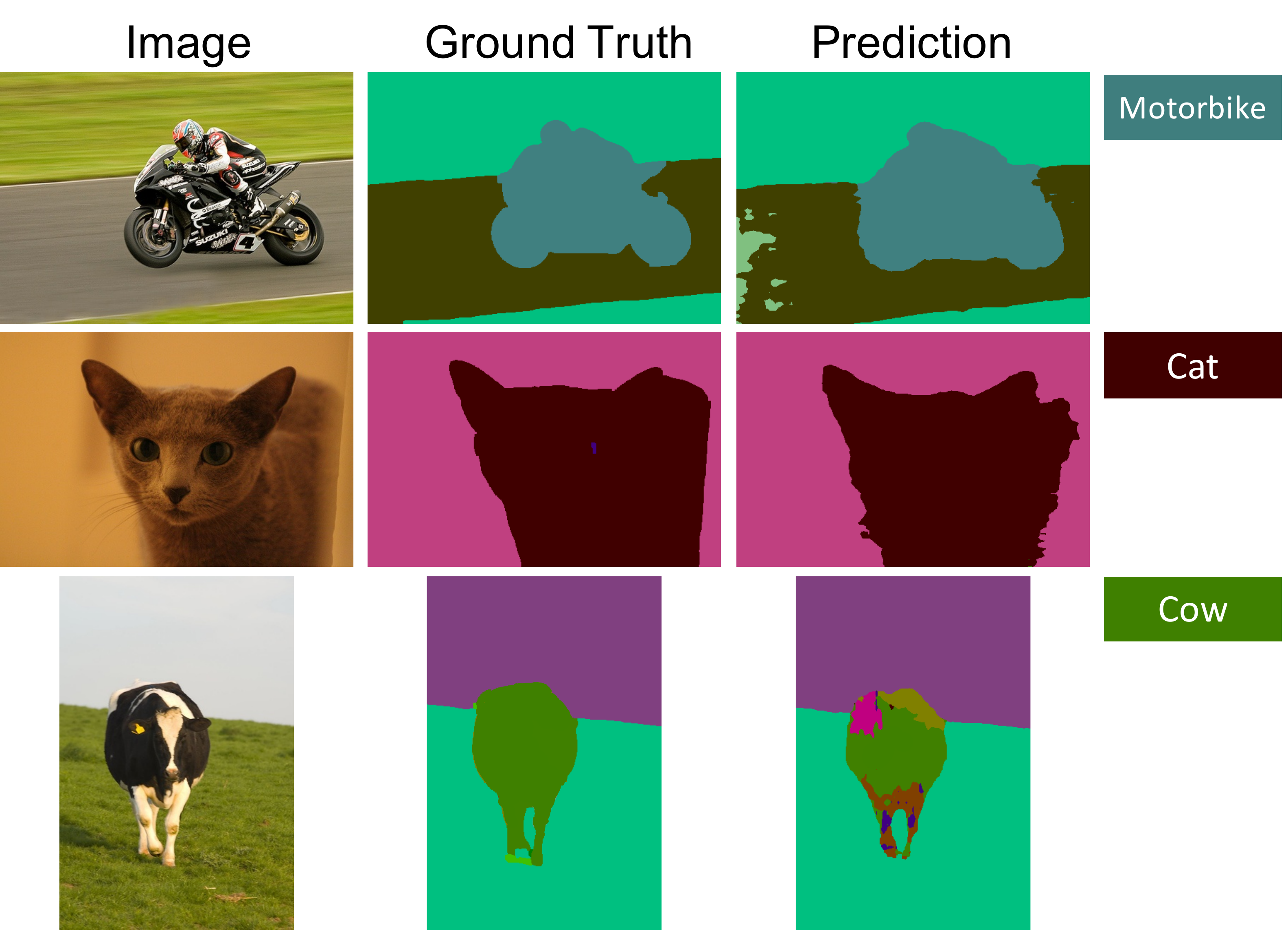}
    \caption{Additional qualitative results on Pascal Context. Rectangle on the right side indicates the unseen class in the sample on the left}
    \label{fig.context_supp}
\end{figure*}

\begin{figure*}
    \centering
    \includegraphics[width=0.8\linewidth]{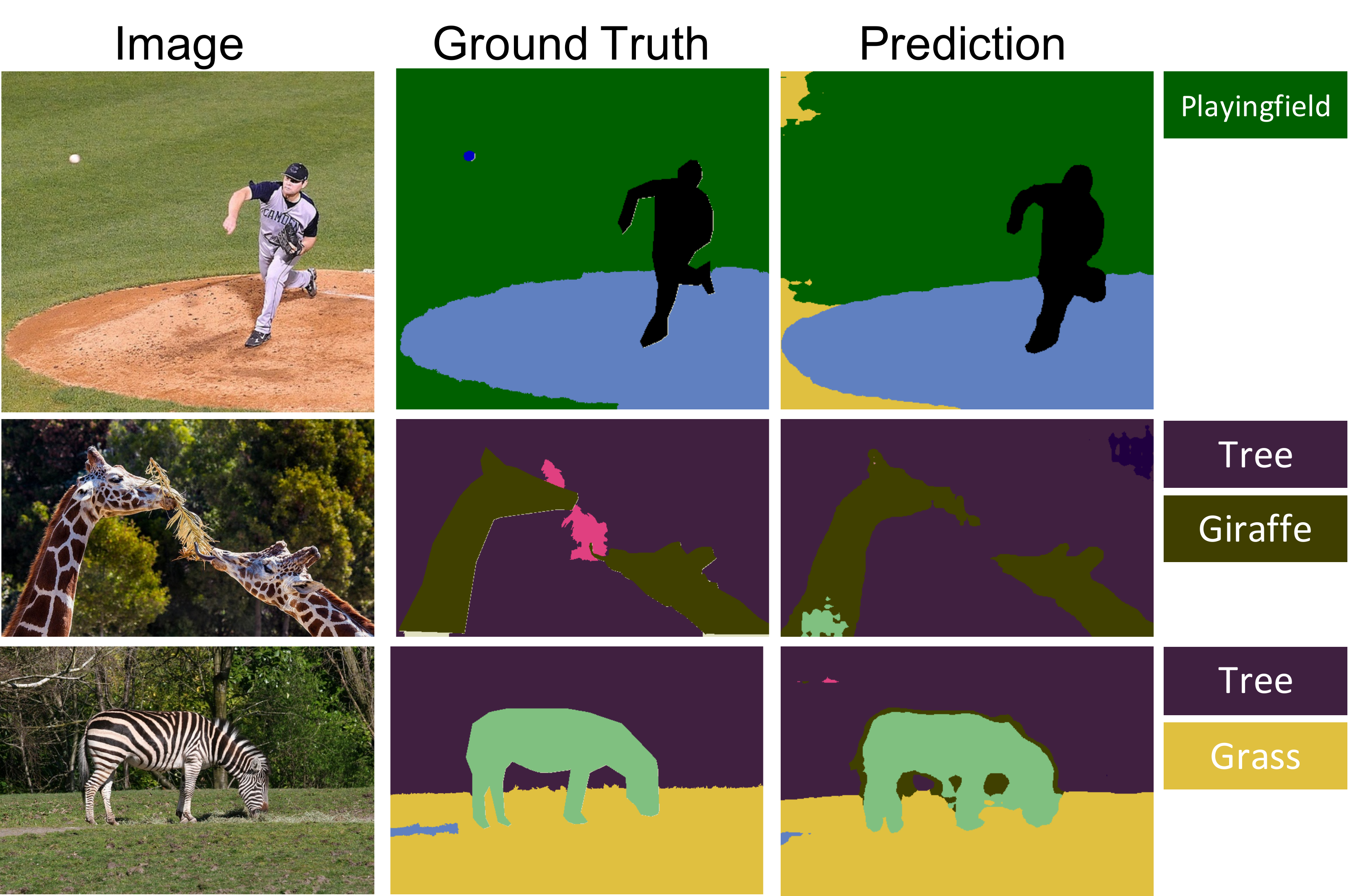}
    \caption{Additional qualitative results on COCO Stuff. Rectangle(s) on the right side indicates the unseen class(es) in the sample on the left}
    \label{fig.coco_supp}
\end{figure*}

We show additional qualitative results of our SIGN model in \cref{fig.voc_supp,fig.context_supp,fig.coco_supp}.

\end{document}